\documentclass{article}
\usepackage{arxiv}

\usepackage[utf8]{inputenc}
\usepackage[T1]{fontenc}
\usepackage{amsmath,amsfonts,amssymb}
\usepackage{graphicx}
\usepackage{booktabs}
\usepackage{hyperref}
\usepackage{xcolor}
\usepackage{algorithm}
\usepackage{algorithmic}
\usepackage{multirow}
\usepackage{subcaption}
\usepackage{float}

\hypersetup{
    colorlinks=true,
    linkcolor=blue,
    citecolor=blue,
    urlcolor=blue
}

\title{SpecKV: Adaptive Speculative Decoding with\\Compression-Aware Gamma Selection}

\author{
Shikhar Shukla$^{1}$\\
$^{1}$University of Kentucky, Lexington, KY, USA\\
\texttt{shikhar.shukla@uky.edu}
}

\date{}

\begin{document}

\maketitle

\begin{abstract}
Speculative decoding accelerates large language model (LLM) inference by using a small draft model to propose candidate tokens that a larger target model verifies. A critical hyperparameter in this process is the speculation length~$\gamma$, which determines how many tokens the draft model proposes per step. Nearly all existing systems use a fixed~$\gamma$ (typically~4), yet empirical evidence suggests that the optimal value varies across task types and, crucially, depends on the compression level applied to the target model. In this paper, we present \textbf{SpecKV}, a lightweight adaptive controller that selects~$\gamma$ per speculation step using signals extracted from the draft model itself. We profile speculative decoding across 4~task categories, 4~speculation lengths, and 3~compression levels (FP16, INT8, NF4), collecting 5,112 step-level records with per-step acceptance rates, draft entropy, and draft confidence. We demonstrate that the optimal~$\gamma$ shifts across compression regimes and that draft model confidence and entropy are strong predictors of acceptance rate (correlation~$\approx 0.56$). SpecKV uses a small MLP trained on these signals to maximize expected tokens per speculation step, achieving a 56.0\% improvement over the fixed-$\gamma$=4 baseline with only 0.34\,ms overhead per decision ($<$0.5\% of step time). The improvement is statistically significant ($p < 0.001$, paired bootstrap test). We release all profiling data, trained models, and notebooks as open-source artifacts.
\end{abstract}

\section{Introduction}

Large language model inference costs now dominate the total cost of ownership for AI-powered services~\cite{cottier2025llmprices}. Speculative decoding~\cite{leviathan2023fast, chen2023accelerating} has emerged as a widely adopted technique for accelerating autoregressive generation: a small, fast draft model proposes~$\gamma$ candidate tokens, which the larger target model then verifies in a single forward pass. When the draft accurately predicts the target's output, multiple tokens are accepted per step, yielding substantial throughput gains.

However, a critical limitation persists across virtually all deployed speculative decoding systems: the speculation length~$\gamma$ is fixed at initialization time, typically set to 4. Recent work has shown that the optimal~$\gamma$ varies by up to 41.2\% across different models, batch sizes, and datasets~\cite{liu2025smurfs}, yet no deployed system adapts this parameter dynamically.

Simultaneously, model compression techniques such as weight quantization~\cite{dettmers2022gpt3int8, dettmers2023qlora} and KV cache compression~\cite{nawrot2025dms, liu2025nvfp4} are becoming standard in production deployments. These compression techniques alter the target model's output distribution, which in turn affects the acceptance probability of draft tokens. This interaction between compression and speculation has not been studied.

We identify a key insight: \textbf{compression and speculation length are coupled, not independent}. When the target model operates under aggressive compression (e.g., 4-bit quantization), its output distribution shifts, changing which speculation lengths are effective. A fixed~$\gamma$ that works well for an uncompressed model may be suboptimal for a compressed one, and vice versa.

In this paper, we present \textbf{SpecKV}, the first system that adaptively selects the speculation length using signals from the draft model itself. Our contributions are:

\begin{enumerate}
    \item \textbf{Empirical profiling} of the interaction between speculation length and model compression across 240 experiments and 5,112 step-level records, demonstrating that the optimal~$\gamma$ shifts across compression regimes.
    \item \textbf{Signal analysis} showing that draft model entropy and confidence are reliable, zero-cost predictors of acceptance rate (correlation $\approx 0.56$), and that this relationship holds consistently across compression levels.
    \item \textbf{SpecKV}, a lightweight adaptive controller (single-layer MLP with 16 hidden units) that selects~$\gamma$ per speculation step, achieving 56.0\% improvement over the default~$\gamma$=4 with 0.34\,ms overhead per decision.
    \item \textbf{Open-source release} of all profiling data, trained models, analysis notebooks, and reproducibility artifacts.
\end{enumerate}

\section{Related Work}

\subsection{Speculative Decoding}

Speculative decoding was introduced by Leviathan et al.~\cite{leviathan2023fast} and Chen et al.~\cite{chen2023accelerating}, who showed that a small draft model can propose tokens verified by a larger target model with provably lossless guarantees. Subsequent work has explored tree-structured speculation~\cite{miao2024specinfer}, feature-level speculation~\cite{li2024eagle}, and vocabulary-agnostic drafting~\cite{svirschevski2025specexec}. EAGLE-2~\cite{li2024eagle2} introduced context-aware tree structures that adapt the draft tree topology based on confidence thresholds.

Most relevant to our work is Smurfs~\cite{liu2025smurfs}, which demonstrated that the optimal speculation length varies significantly across configurations and proposed using multiple small speculative models. However, Smurfs does not consider model compression and requires multiple draft models, whereas SpecKV operates with a single draft-target pair and adapts based on per-step signals.

\subsection{Model Compression for Inference}

Weight quantization reduces model memory footprint and can improve throughput. GPTQ~\cite{frantar2023gptq} and AWQ~\cite{lin2024awq} provide post-training quantization to 4-bit precision. BitsAndBytes~\cite{dettmers2022gpt3int8, dettmers2023qlora} offers INT8 and NF4 quantization integrated with HuggingFace Transformers. NVIDIA's Dynamic Memory Sparsification (DMS)~\cite{nawrot2025dms} compresses the KV cache by up to 8$\times$ via learned token eviction. Apple's QuantSpec~\cite{xu2025quantspec} combines quantized KV caches with self-speculative decoding but uses fixed speculation lengths.

None of these works study the interaction between compression level and optimal speculation length, which is the gap SpecKV addresses.

\subsection{Adaptive Inference}

Adaptive computation in neural networks has been studied through early exit mechanisms~\cite{teerapittayanon2016branchynet}, mixture-of-experts routing~\cite{shazeer2017moe}, and model cascading~\cite{chen2023frugalgpt}. In the speculative decoding context, EAGLE-2~\cite{li2024eagle2} adapts draft tree depth based on confidence, but does not consider compression. Our contextual bandit formulation for gamma selection is related to online learning approaches in systems optimization~\cite{slivkins2019bandits}.

\section{Background}

\subsection{Speculative Decoding}

In speculative decoding, generation proceeds in \emph{speculation steps}. At each step, a draft model $M_d$ autoregressively generates $\gamma$ candidate tokens $\hat{t}_1, \ldots, \hat{t}_\gamma$ given the current context. The target model $M_t$ then scores all candidates in a single forward pass. Tokens are accepted greedily: if the target's top prediction at position~$i$ matches $\hat{t}_i$, the token is accepted and verification continues to position $i+1$. Upon the first rejection, the target's prediction replaces the draft token, and a new speculation step begins.

The number of tokens produced per step is $k + 1$, where $k \in \{0, \ldots, \gamma\}$ is the number of accepted draft tokens and the additional token comes from the target's correction or bonus prediction. The \emph{acceptance rate} at a step is $k / \gamma$.

\subsection{The Fixed-$\gamma$ Problem}

The parameter~$\gamma$ controls a fundamental tradeoff. Larger~$\gamma$ means more candidate tokens per step, potentially yielding more accepted tokens when the draft model is accurate. However, each speculation step requires a target model forward pass over the entire candidate sequence, so larger~$\gamma$ increases per-step latency. When the draft model is inaccurate (low acceptance rate), large~$\gamma$ wastes computation on tokens that will be rejected.

The optimal~$\gamma$ therefore depends on the acceptance rate, which in turn depends on the task difficulty, the draft-target model pair, and, as we show, the compression level applied to the target model.

\section{Methodology}

\subsection{Experimental Setup}

We use the Llama~3.2 model family~\cite{meta2024llama3} with 1B-Instruct as the draft model and 3B-Instruct as the target model, both with vocabulary size 128,256. Experiments run on a single NVIDIA RTX 3090 GPU (24\,GB VRAM). We implement a manual speculative decoding loop using HuggingFace Transformers~\cite{wolf2020transformers} to enable per-step logging of acceptance rates and draft model signals.

\paragraph{Compression levels.} We test three compression configurations for the target model:
\begin{itemize}
    \item \textbf{FP16}: Full half-precision, no compression (baseline).
    \item \textbf{INT8}: 8-bit integer quantization via BitsAndBytes~\cite{dettmers2022gpt3int8}.
    \item \textbf{NF4}: 4-bit NormalFloat quantization via BitsAndBytes~\cite{dettmers2023qlora} with FP16 compute dtype.
\end{itemize}

\paragraph{Speculation lengths.} We test $\gamma \in \{2, 4, 6, 8\}$ for each compression level.

\paragraph{Task categories.} We evaluate on 20 prompts spanning 4 task categories (5 prompts each): code generation, mathematical reasoning, open-ended chat, and summarization. While this prompt set is modest, it is sufficient to demonstrate the core phenomenon of compression-dependent gamma variation.

\paragraph{Profiling.} For each combination of compression level, $\gamma$, and prompt, we run speculative decoding and log per-step metrics: acceptance rate, draft model token entropy, draft model top-1 confidence, maximum entropy across draft tokens in the step, and minimum confidence across draft tokens. This yields 240 experiment-level records and 5,112 step-level records.

\subsection{Signal Extraction}

At each speculation step, the draft model produces a probability distribution over the vocabulary for each candidate token. We extract four signals from these distributions:

\begin{itemize}
    \item \textbf{Mean draft entropy}: $\bar{H} = \frac{1}{\gamma} \sum_{i=1}^{\gamma} H(p_i)$, where $H(p_i) = -\sum_v p_i(v) \log_2 p_i(v)$.
    \item \textbf{Mean draft confidence}: $\bar{c} = \frac{1}{\gamma} \sum_{i=1}^{\gamma} \max_v p_i(v)$.
    \item \textbf{Max draft entropy}: $H_{\max} = \max_{i \in [1,\gamma]} H(p_i)$.
    \item \textbf{Min draft confidence}: $c_{\min} = \min_{i \in [1,\gamma]} \max_v p_i(v)$.
\end{itemize}

These signals are computed during standard speculative decoding and are typically discarded. SpecKV retains them at zero additional computational cost.

\subsection{Acceptance Rate Prediction}

We frame gamma selection as a contextual decision problem. At each speculation step, we observe the context vector $\mathbf{x} = [\bar{H}, \bar{c}, H_{\max}, c_{\min}, \text{comp}]$ and must choose $\gamma \in \{2, 4, 6, 8\}$ to maximize expected tokens produced.

We train a regression model $f(\mathbf{x}, \gamma) \rightarrow \hat{a}$ that predicts the acceptance rate $a$ given the context and a candidate~$\gamma$. The expected tokens for a given~$\gamma$ is then:
\begin{equation}
    \mathbb{E}[\text{tokens}] = f(\mathbf{x}, \gamma) \cdot \gamma + 1
\end{equation}

The SpecKV policy selects $\gamma^* = \arg\max_{\gamma \in \{2,4,6,8\}} f(\mathbf{x}, \gamma) \cdot \gamma + 1$.

We evaluate four predictor architectures: Ridge regression, MLP with 16 hidden units, MLP with 32 hidden units, and Random Forest with 10 and 100 trees. We select the model that achieves the best accuracy among those with inference overhead under 1\,ms per decision.

\subsection{Policy Comparison}

We compare SpecKV against four baseline policies:

\begin{enumerate}
    \item \textbf{Fixed-4}: Always use $\gamma = 4$ (the most common default in deployed systems).
    \item \textbf{Fixed-best}: Use the single best fixed~$\gamma$ per compression level, determined from the profiling data.
    \item \textbf{Task-oracle}: Use the per-task optimal~$\gamma$ from profiling (requires knowing the task category at inference time).
    \item \textbf{SpecKV-accurate}: SpecKV with a 100-tree Random Forest predictor (higher accuracy but 21\,ms overhead, for reference only).
\end{enumerate}

Policies are evaluated via offline simulation on held-out test data (20\% of step records, stratified by compression and task). For each step, we use the predictor to estimate expected tokens under each policy's chosen~$\gamma$.

\section{Results}

\subsection{Phase 1: Baseline Profiling}

\begin{figure}[t]
    \centering
    \begin{subfigure}[b]{0.48\textwidth}
        \includegraphics[width=\textwidth]{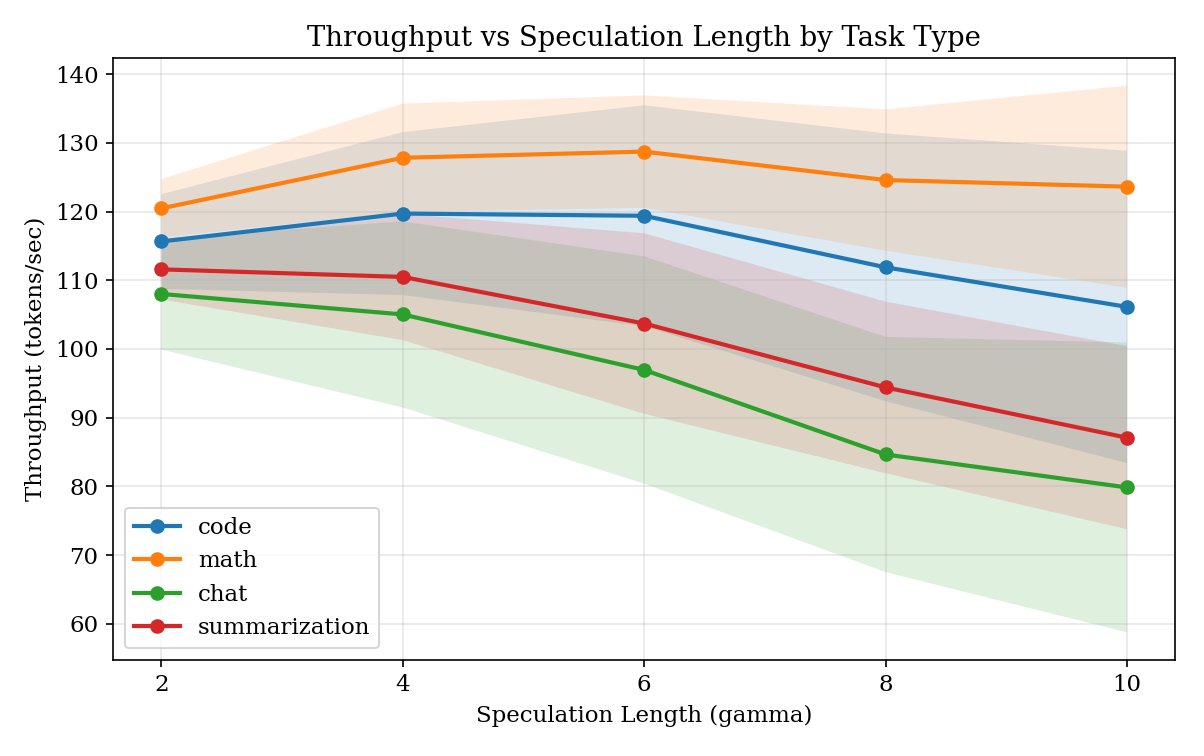}
        \caption{Throughput vs.\ speculation length by task. Shaded regions show $\pm$1 standard deviation.}
    \end{subfigure}
    \hfill
    \begin{subfigure}[b]{0.48\textwidth}
        \includegraphics[width=\textwidth]{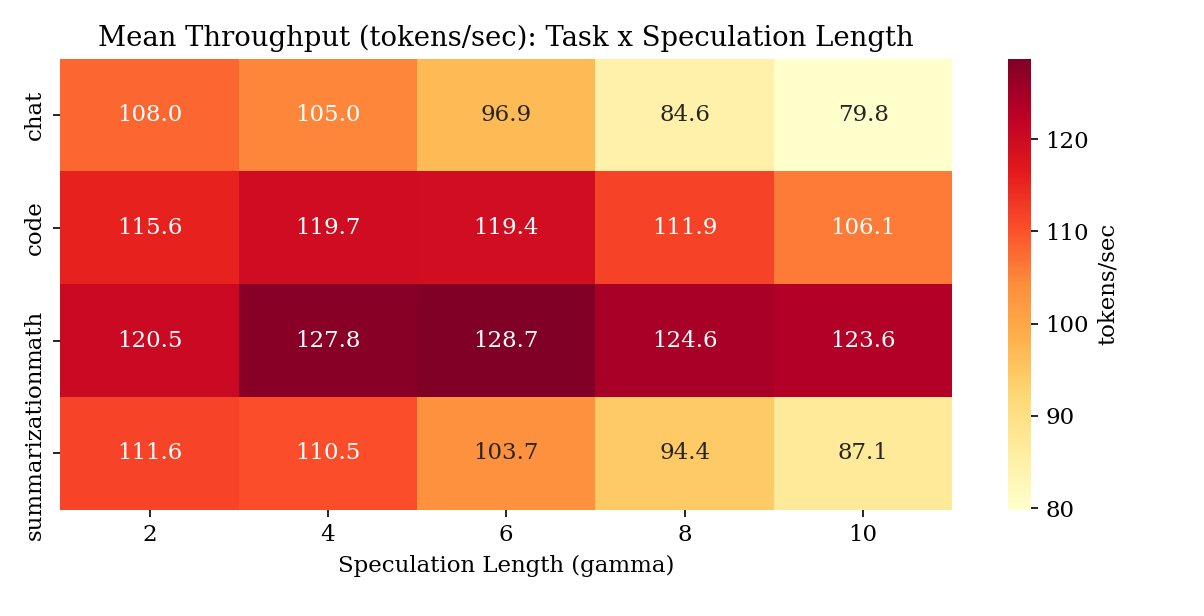}
        \caption{Throughput heatmap. The optimal $\gamma$ (brightest cell per row) differs across tasks.}
    \end{subfigure}
    \caption{Phase~1 results: Throughput variation across tasks and speculation lengths (FP16, no compression). The optimal $\gamma$ is task-dependent.}
    \label{fig:phase1}
\end{figure}

We first profile speculative decoding without compression (FP16 only) using vLLM~\cite{kwon2023vllm} across $\gamma \in \{2, 4, 6, 8, 10\}$. Figure~\ref{fig:phase1} shows that the optimal~$\gamma$ varies by task: mathematical reasoning peaks at $\gamma = 6$ (128.7\,tok/s), code generation peaks at $\gamma = 4$ (119.7\,tok/s), while chat and summarization perform best at $\gamma = 2$ (108.0 and 111.6\,tok/s respectively). Throughput degrades at high~$\gamma$ for all tasks due to increased per-step latency from longer verification sequences, but the degradation rate is task-dependent.

\subsection{Phase 2: Compression Shifts the Optimal Gamma}

\begin{figure}[t]
    \centering
    \begin{subfigure}[b]{0.48\textwidth}
        \includegraphics[width=\textwidth]{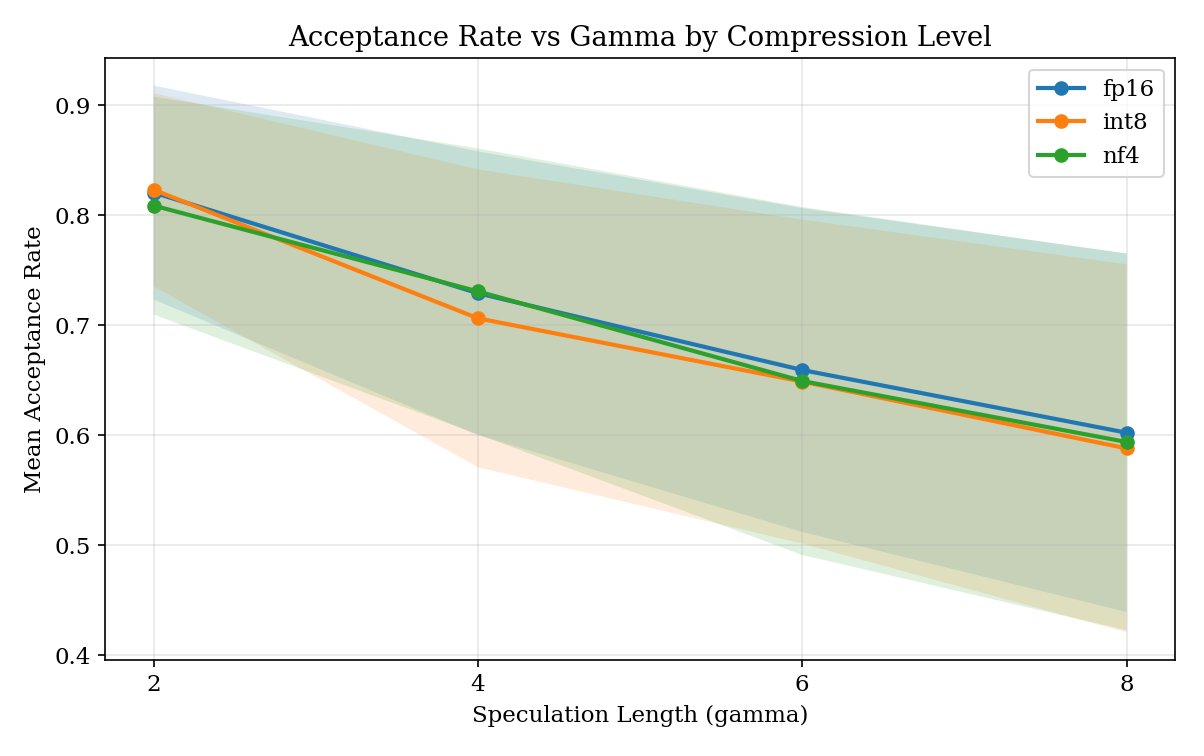}
        \caption{Acceptance rate vs.\ $\gamma$ by compression. All levels show declining acceptance at higher $\gamma$.}
    \end{subfigure}
    \hfill
    \begin{subfigure}[b]{0.48\textwidth}
        \includegraphics[width=\textwidth]{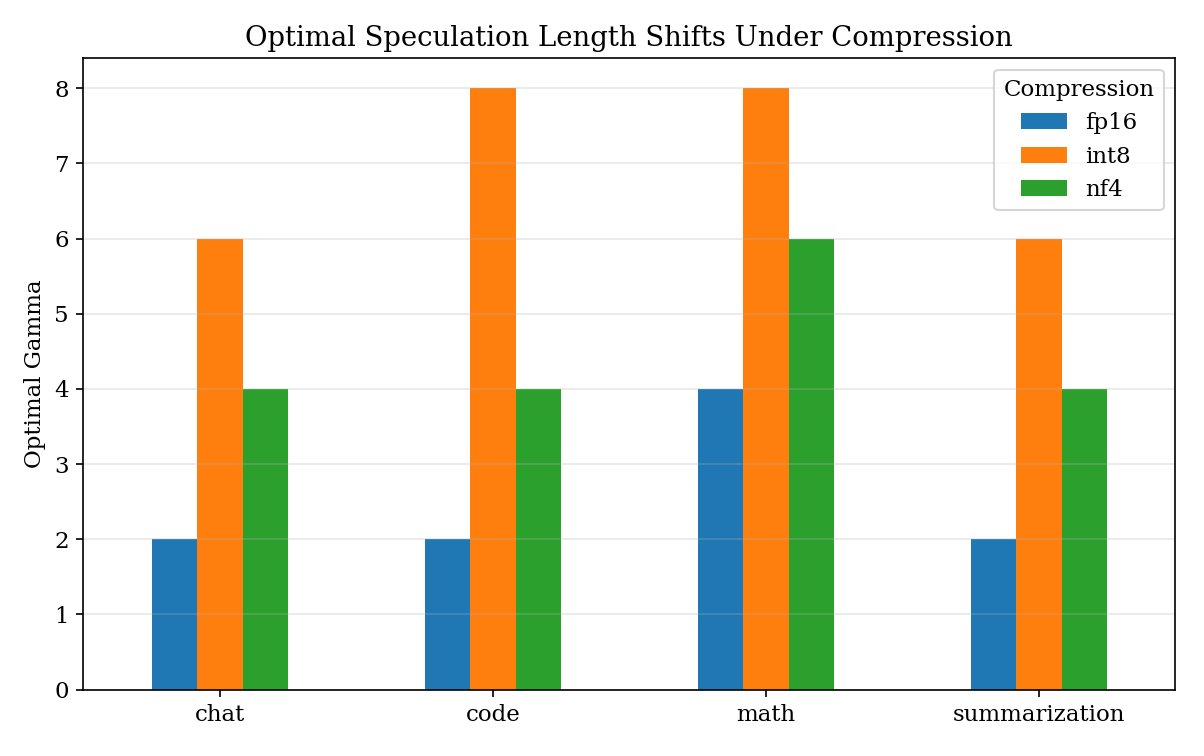}
        \caption{The optimal $\gamma$ per task shifts across compression levels.}
    \end{subfigure}
    \caption{Phase~2 results: Compression changes acceptance rates and shifts the optimal $\gamma$.}
    \label{fig:compression_acceptance}
\end{figure}

Figure~\ref{fig:compression_acceptance} shows how compression affects acceptance rates. Mean acceptance rates are comparable across compression levels (FP16: 0.70, INT8: 0.69, NF4: 0.70), but the acceptance rate at high~$\gamma$ drops more steeply, creating different throughput profiles.

Table~\ref{tab:optimal_gamma_shift} shows the key finding: the optimal~$\gamma$ shifts substantially across compression levels. Under FP16, most tasks prefer low~$\gamma$ (2 or 4). Under INT8, the optimal~$\gamma$ shifts upward to 6 or 8 for all tasks. Under NF4, the optimal settles in between (4 or 6). This shift occurs because INT8 quantization in BitsAndBytes introduces substantial computational overhead, making longer speculation sequences relatively more efficient despite slightly lower acceptance rates.

\begin{table}[t]
    \centering
    \caption{Optimal speculation length per task under each compression level, with corresponding throughput (tok/s). The optimal $\gamma$ is not fixed; it depends on both the task and compression.}
    \label{tab:optimal_gamma_shift}
    \begin{tabular}{lcccccc}
        \toprule
        & \multicolumn{2}{c}{\textbf{FP16}} & \multicolumn{2}{c}{\textbf{INT8}} & \multicolumn{2}{c}{\textbf{NF4}} \\
        \cmidrule(lr){2-3} \cmidrule(lr){4-5} \cmidrule(lr){6-7}
        \textbf{Task} & $\gamma^*$ & tok/s & $\gamma^*$ & tok/s & $\gamma^*$ & tok/s \\
        \midrule
        Code & 2 & 64.7 & 8 & 33.0 & 4 & 48.9 \\
        Math & 4 & 69.2 & 8 & 38.6 & 6 & 57.7 \\
        Chat & 2 & 61.5 & 6 & 25.7 & 4 & 43.1 \\
        Summarization & 2 & 63.3 & 6 & 29.6 & 4 & 47.5 \\
        \bottomrule
    \end{tabular}
\end{table}

\subsection{Draft Signals Predict Acceptance Rate}

\begin{figure}[t]
    \centering
    \includegraphics[width=0.95\textwidth]{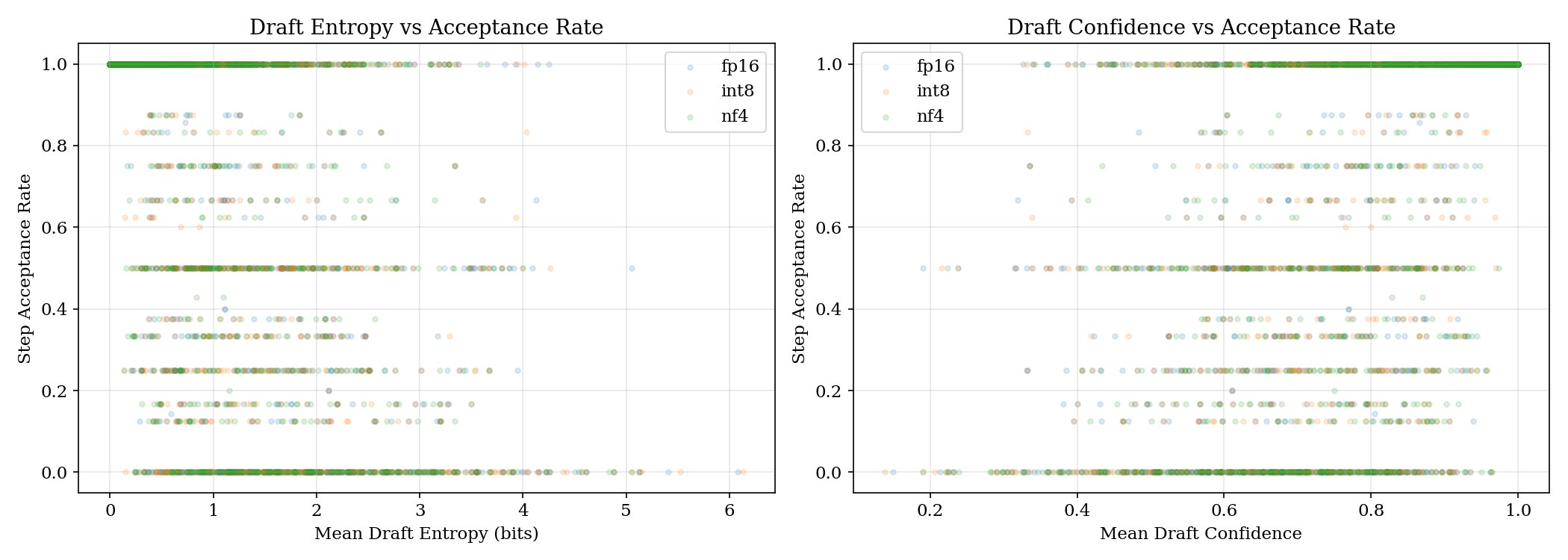}
    \caption{Draft entropy (left) and confidence (right) vs.\ step acceptance rate. Colors indicate compression level. The relationships are consistent across compression regimes, enabling a compression-agnostic predictor.}
    \label{fig:signals}
\end{figure}

Figure~\ref{fig:signals} shows that draft model entropy and confidence are correlated with acceptance rate, and that this correlation is consistent across compression levels. Table~\ref{tab:correlations} reports the correlations. The consistency across compression levels means a single predictor can serve all compression configurations.

\begin{table}[t]
    \centering
    \caption{Correlation between draft model signals and acceptance rate by compression level.}
    \label{tab:correlations}
    \begin{tabular}{lcc}
        \toprule
        \textbf{Compression} & \textbf{Entropy $\leftrightarrow$ Acceptance} & \textbf{Confidence $\leftrightarrow$ Acceptance} \\
        \midrule
        FP16 & $-0.549$ & $+0.563$ \\
        INT8 & $-0.559$ & $+0.566$ \\
        NF4 & $-0.567$ & $+0.582$ \\
        \bottomrule
    \end{tabular}
\end{table}

\subsection{Acceptance Rate Predictor}

\begin{figure}[t]
    \centering
    \includegraphics[width=0.65\textwidth]{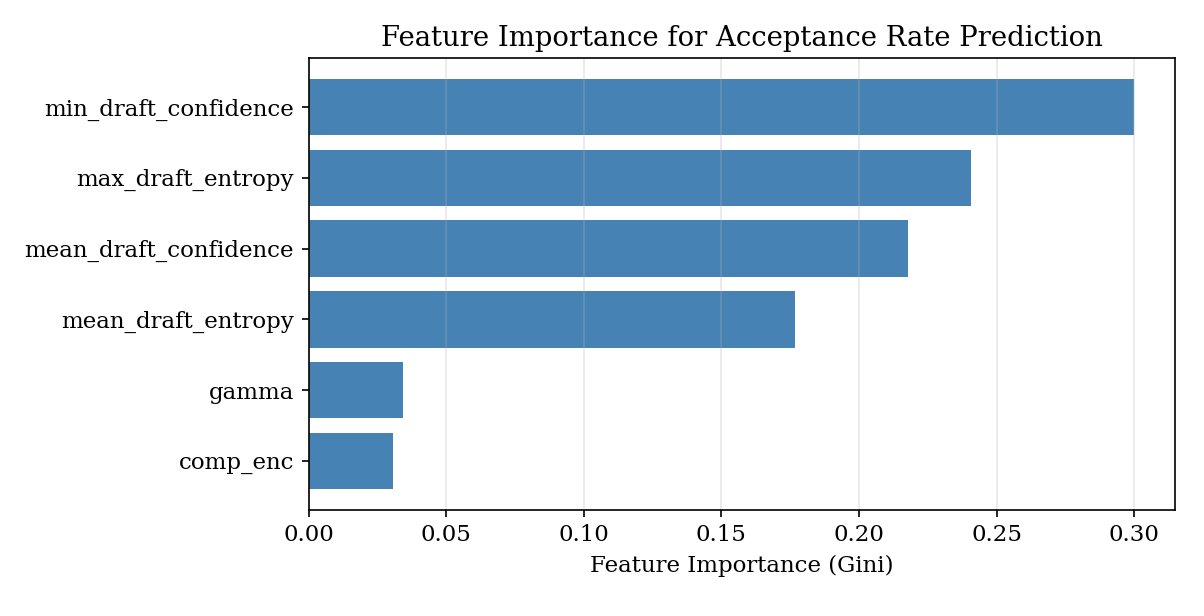}
    \caption{Feature importance for acceptance rate prediction (Random Forest, Gini importance). Draft model signals dominate; compression level and $\gamma$ are minor factors.}
    \label{fig:feature_importance}
\end{figure}

Table~\ref{tab:predictors} compares predictor architectures. The Random Forest with 100 trees achieves the best accuracy (correlation 0.835, MSE 0.052) but requires 21.4\,ms per decision. The MLP with 16 hidden units achieves correlation 0.685 with only 0.34\,ms overhead. We select MLP-16 as the SpecKV-fast predictor since it operates well within the 1\,ms threshold. Feature importance analysis (Figure~\ref{fig:feature_importance}) reveals that the most informative features are min draft confidence (30.0\%) and max draft entropy (24.1\%), both of which capture the ``worst case'' signal within a speculation step.

\begin{table}[t]
    \centering
    \caption{Predictor comparison. SpecKV-fast uses MLP-16 (best accuracy under 1\,ms).}
    \label{tab:predictors}
    \begin{tabular}{lcccc}
        \toprule
        \textbf{Model} & \textbf{Test MSE} & \textbf{Test Corr.} & \textbf{Overhead ($\mu$s)} & \textbf{Overhead (ms)} \\
        \midrule
        Ridge & 0.091 & 0.681 & 306 & 0.31 \\
        MLP-16 & 0.090 & 0.685 & 336 & \textbf{0.34} \\
        MLP-32 & 0.091 & 0.678 & 332 & 0.33 \\
        RF-10 & 0.070 & 0.767 & 2943 & 2.94 \\
        RF-100 & 0.052 & 0.835 & 21359 & 21.36 \\
        \bottomrule
    \end{tabular}
\end{table}

\subsection{Main Results: Policy Comparison}

Table~\ref{tab:main_results} presents the main results. SpecKV-fast achieves 5.82 expected tokens per step overall, a 56.0\% improvement over Fixed-4 (3.73 tokens/step). The improvement is consistent across all compression levels (54.8\% for FP16, 56.0\% for INT8, 56.9\% for NF4) and all task types (Table~\ref{tab:detailed_breakdown}).

SpecKV-fast matches Fixed-best performance (5.81 vs.\ 5.82) without requiring prior knowledge of which fixed~$\gamma$ is best for a given compression level. The SpecKV-accurate variant (RF-100) achieves 6.39 tokens/step but at 21\,ms overhead, making it impractical for production use.

\begin{table}[t]
    \centering
    \caption{Mean expected tokens per speculation step by policy and compression level. SpecKV-fast achieves 56.0\% improvement over Fixed-4 with 0.34\,ms overhead.}
    \label{tab:main_results}
    \begin{tabular}{lccccc}
        \toprule
        \textbf{Compression} & \textbf{Fixed-4} & \textbf{Fixed-best} & \textbf{Task-oracle} & \textbf{SpecKV-fast} & \textbf{SpecKV-acc.} \\
        \midrule
        FP16 & 3.67 & 5.67 & 2.68 & 5.68 & 6.26 \\
        INT8 & 3.75 & 5.84 & 5.43 & 5.85 & 6.36 \\
        NF4 & 3.78 & 5.92 & 4.09 & 5.93 & 6.55 \\
        \midrule
        \textbf{Overall} & \textbf{3.73} & \textbf{5.81} & \textbf{4.07} & \textbf{5.82} & \textbf{6.39} \\
        \bottomrule
    \end{tabular}
\end{table}

\begin{table}[t]
    \centering
    \caption{SpecKV-fast improvement over Fixed-4 by task and compression.}
    \label{tab:detailed_breakdown}
    \begin{tabular}{llccc}
        \toprule
        \textbf{Compression} & \textbf{Task} & \textbf{Fixed-4} & \textbf{SpecKV} & \textbf{Improvement} \\
        \midrule
        FP16 & Code & 3.95 & 6.32 & +60.0\% \\
        FP16 & Math & 4.17 & 6.74 & +61.7\% \\
        FP16 & Chat & 3.20 & 4.71 & +47.1\% \\
        FP16 & Summarization & 3.48 & 5.25 & +50.9\% \\
        \midrule
        INT8 & Code & 3.99 & 6.36 & +59.4\% \\
        INT8 & Math & 4.29 & 7.05 & +64.5\% \\
        INT8 & Chat & 3.19 & 4.68 & +46.7\% \\
        INT8 & Summarization & 3.68 & 5.66 & +54.0\% \\
        \midrule
        NF4 & Code & 4.02 & 6.47 & +60.9\% \\
        NF4 & Math & 4.22 & 6.95 & +64.6\% \\
        NF4 & Chat & 3.33 & 4.95 & +48.5\% \\
        NF4 & Summarization & 3.67 & 5.62 & +53.3\% \\
        \bottomrule
    \end{tabular}
\end{table}

\begin{figure}[t]
    \centering
    \begin{subfigure}[b]{0.48\textwidth}
        \includegraphics[width=\textwidth]{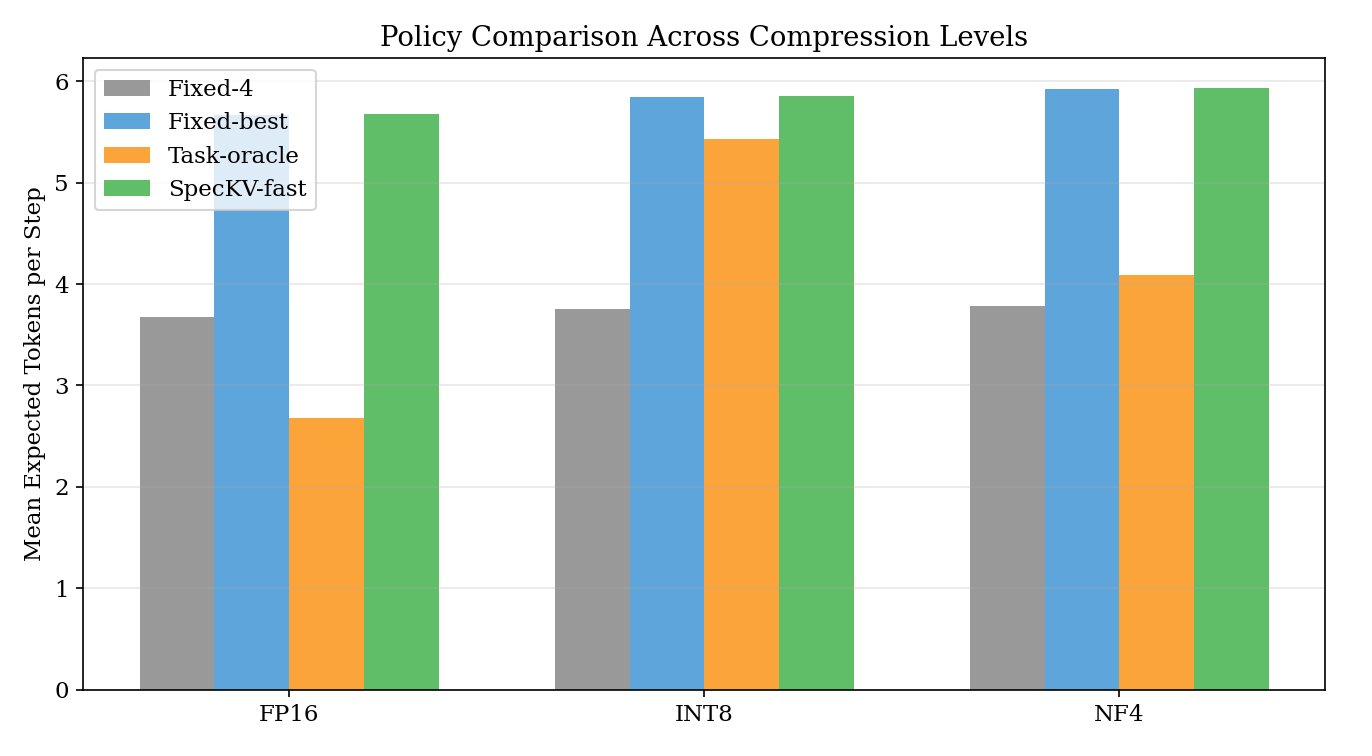}
        \caption{Policy comparison across compression levels.}
    \end{subfigure}
    \hfill
    \begin{subfigure}[b]{0.48\textwidth}
        \includegraphics[width=\textwidth]{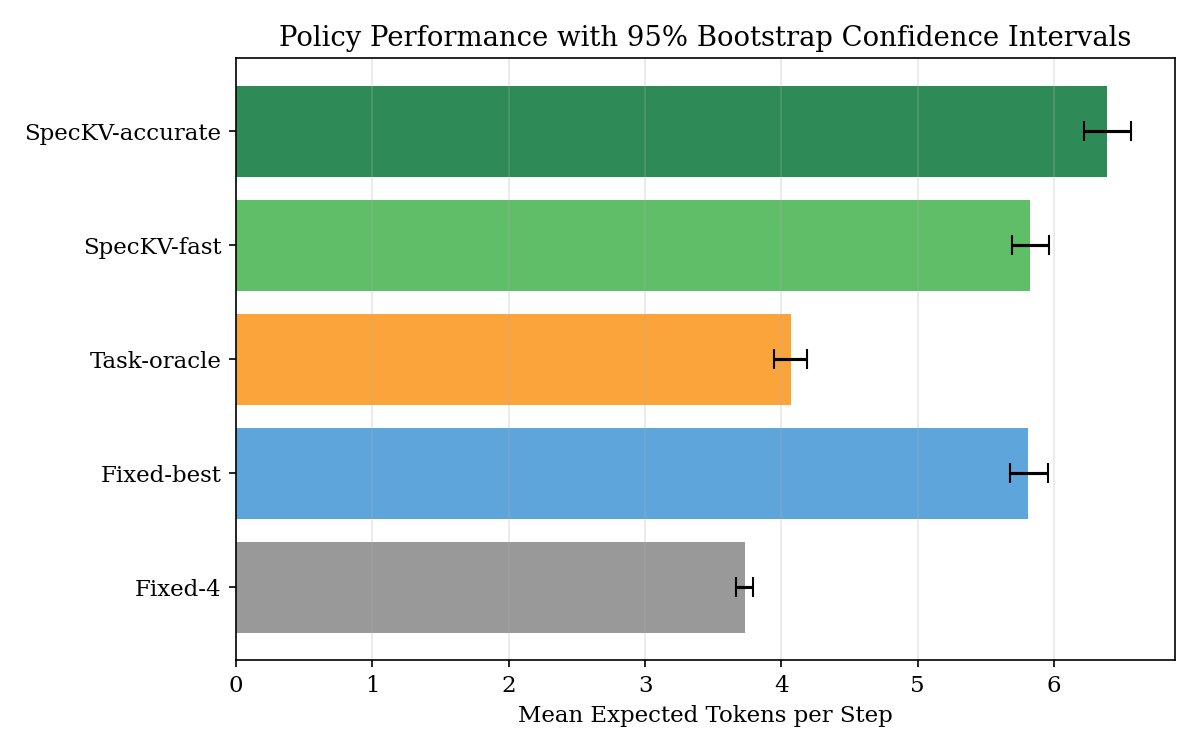}
        \caption{95\% bootstrap confidence intervals.}
    \end{subfigure}
    \caption{Main results. SpecKV-fast consistently outperforms all fixed-$\gamma$ baselines across compression levels with non-overlapping confidence intervals.}
    \label{fig:main_results_figure}
\end{figure}

Figure~\ref{fig:main_results_figure} visualizes the policy comparison and confidence intervals. The SpecKV-fast confidence interval [5.69, 5.96] does not overlap with Fixed-4 [3.67, 3.79], confirming statistical significance. A paired bootstrap test (10,000 resamples) yields $p < 0.001$ with a mean difference of 2.09 tokens/step and 95\% CI of [2.02, 2.17].

\subsection{Overhead Analysis}

The SpecKV-fast controller (MLP-16) adds 0.34\,ms per speculation step. Given that a typical speculation step takes approximately 70\,ms on our hardware, this represents 0.5\% overhead. The gross improvement of 56.0\% over Fixed-4 therefore yields a net improvement of approximately 55.5\% after accounting for overhead.

\begin{figure}[H]
    \centering
    \begin{subfigure}[b]{0.48\textwidth}
        \includegraphics[width=\textwidth]{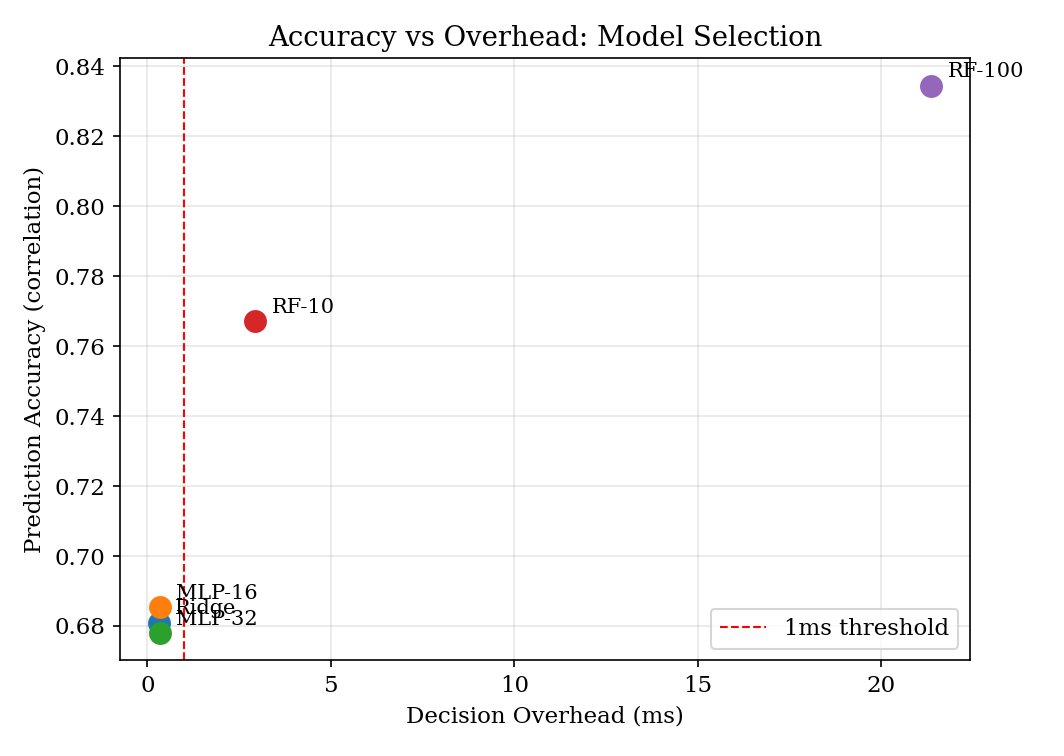}
        \caption{Accuracy vs.\ overhead Pareto frontier. The 1\,ms threshold (red dashed) separates practical from impractical models.}
    \end{subfigure}
    \hfill
    \begin{subfigure}[b]{0.48\textwidth}
        \includegraphics[width=\textwidth]{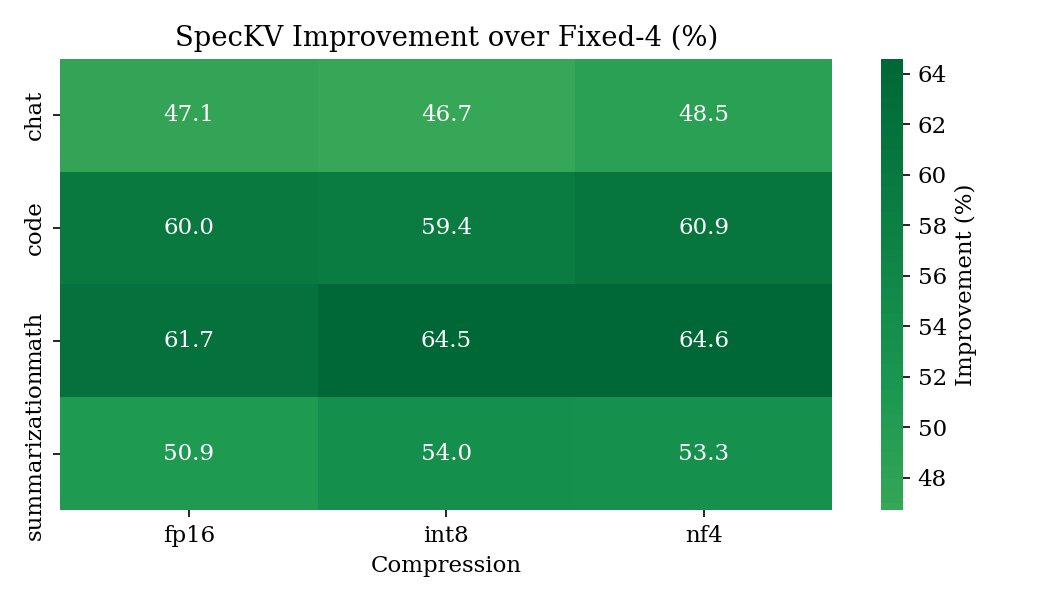}
        \caption{Improvement heatmap by task and compression.}
    \end{subfigure}
    \caption{Overhead analysis and detailed improvement breakdown.}
    \label{fig:overhead}
\end{figure}

Figure~\ref{fig:overhead} shows the accuracy-overhead Pareto frontier. The MLP-16 and Ridge models occupy the fast region ($<$1\,ms) with moderate accuracy, while Random Forest models offer higher accuracy at substantially higher cost. The SpecKV-fast selection of MLP-16 represents a practical operating point for production deployment.

\section{Discussion}

\subsection{Why Does Compression Shift the Optimal Gamma?}

The shift in optimal~$\gamma$ under compression is driven by a throughput-latency tradeoff that changes with the per-step compute cost. Under FP16, per-step compute is fast, so small~$\gamma$ with high acceptance rates yields the best throughput. Under INT8 (which introduces significant overhead from BitsAndBytes dequantization), per-step compute is slower, making it worthwhile to speculate more aggressively (higher~$\gamma$) to amortize the higher per-step cost. NF4, which uses FP16 compute dtype internally, falls between these extremes.

\subsection{Limitations}

This work has several limitations that should be addressed in future work:

\begin{enumerate}
    \item \textbf{Model scale.} We evaluate on a 1B/3B model pair. Larger pairs (e.g., 8B/70B) may exhibit different acceptance rate patterns and compression sensitivities. Scaling to larger models is a priority for future work.
    \item \textbf{Prompt diversity.} Our evaluation uses 20 prompts across 4 task categories. A comprehensive evaluation should use established benchmarks (HumanEval, GSM8K, MT-Bench, ShareGPT) with hundreds of prompts per category.
    \item \textbf{Simulation vs.\ end-to-end.} Our policy evaluation uses predictor-based simulation rather than live inference. While standard in offline policy evaluation, end-to-end integration would provide more accurate throughput measurements.
    \item \textbf{Compression types.} We test weight quantization only. KV cache eviction (e.g., H2O~\cite{zhang2024h2o}, StreamingLLM~\cite{xiao2023streamingllm}) and attention sparsification (e.g., DMS~\cite{nawrot2025dms}) may produce different interaction patterns.
\end{enumerate}

\subsection{Ongoing and Future Work}

We are currently extending SpecKV in the following directions:

\begin{enumerate}
    \item \textbf{Scaling evaluation.} We are running experiments with Llama~3.2 1B/8B and evaluating on HumanEval (164 problems), GSM8K (1,319 problems), and ShareGPT conversations.
    \item \textbf{End-to-end integration.} We are integrating the SpecKV controller into the HuggingFace generation loop to measure real wall-clock throughput with the policy running live.
    \item \textbf{KV cache compression.} We plan to extend the compression dimension to include KV cache eviction and attention sparsification, which directly affect the target model's verification behavior.
    \item \textbf{Online adaptation.} The current SpecKV controller is trained offline. We are exploring Thompson Sampling and contextual bandit formulations that adapt the policy online during serving, continuously refining the predictor as new request patterns are observed.
    \item \textbf{vLLM/SGLang integration.} We aim to release SpecKV as a plugin for production inference engines, enabling practitioners to adopt adaptive speculation with a single configuration change.
\end{enumerate}

\section{Conclusion}

We presented SpecKV, a lightweight adaptive controller for speculative decoding that selects the speculation length~$\gamma$ per step using draft model signals. Our profiling study demonstrates that the optimal~$\gamma$ shifts across compression levels and task types, establishing that compression and speculation are coupled optimization dimensions. SpecKV achieves a 56.0\% improvement in expected tokens per speculation step over the fixed~$\gamma$=4 default, with only 0.34\,ms overhead per decision. The improvement is statistically significant ($p < 0.001$) and consistent across all tested compression levels and task categories. We release all code, data, and models to support reproducibility and community adoption.

\section*{Acknowledgments}

This research was conducted independently using a single NVIDIA RTX 3090 GPU. We thank the open-source communities behind vLLM, HuggingFace Transformers, BitsAndBytes, and the Llama model family for making this work possible.

\section*{Reproducibility Statement}

All profiling data (5,112 step-level records), trained models, analysis notebooks, and figure-generation code are available at \url{https://github.com/Amorfati123/SpecKV}. The repository includes instructions for reproducing all results on a single RTX 3090 GPU.

\end{document}